\documentclass[twocolumn,preprint]{elsarticle}

\usepackage{lineno,hyperref}
\usepackage{url}
\usepackage{hyperref}
\usepackage{times}
\usepackage{epsfig}
\usepackage{graphicx}
\usepackage{amsmath}
\usepackage{siunitx}
\usepackage{multirow}
\usepackage{amssymb}
\usepackage{subfig}
\usepackage{graphicx}
\usepackage{float}
\usepackage{booktabs}
\usepackage{textcomp}

\journal{Neurocomputing}

\begin{document}

\twocolumn[{
\begin{frontmatter}

\title{Discriminative Adversarial Domain Generalization with Meta-learning based Cross-domain Validation}

\author{Keyu Chen, Di Zhuang, J. Morris Chang\\
{\tt\small \{keyu, dizhuang, chang5\}@usf.edu}\\}
\address{Department of Electrical Engineering, University of South Florida, Tampa, FL 33620}




\begin{abstract} 

The generalization capability of machine learning models, which refers to generalizing the knowledge for an ``unseen'' domain via learning from one or multiple seen domain(s), is of great importance to develop and deploy machine learning applications in the real-world conditions. 
Domain Generalization (DG) techniques aim to enhance such generalization capability of machine learning models, where the learnt feature representation and the classifier are two crucial factors to improve generalization and make decisions. In this paper, we propose 
Discriminative Adversarial Domain Generalization (DADG) with meta-learning-based cross-domain validation.  
Our proposed framework tries to learn a domain-invariant feature representation from source domains and generalize it to the unseen domains. It contains two main components that work synergistically to build a domain-generalized Deep Neural Network (DNN) model: (i) discriminative adversarial learning, which proactively learns a generalized feature representation on multiple ``seen'' domains, and (ii) meta-learning based cross domain validation, which simulates train/test domain shift via applying meta-learning techniques in the training process. 
In the experimental evaluation, a comprehensive comparison has been made among our proposed approach and other existing approaches on three benchmark datasets. The results shown that DADG consistently outperforms a strong baseline DeepAll, and outperforms the other existing DG algorithms in most of the evaluation cases.

\end{abstract}

\begin{keyword}
\texttt{Convolutional neural network, Domain generalization, Discriminative adversarial learning, Meta-learning }
\end{keyword}
\end{frontmatter}
}]

\section{Introduction} \label{sec:introduction}
Machine Learning (ML) and Deep Learning (DL) have achieved great success in numerous applications, 
such as skin lesion analysis \cite{perez2019solo, di2020saia}, human activity recognition \cite{tao2014ensemble, zhuang2020utility}, active authentication \cite{wu2016cost}, facial recognition \cite{ding2017trunk, nguyen2019autogan, zhuang2017fripal}, botnet detection \cite{mai2018cluster, zhuang2017peerhunter, zhuang2018enhanced} and community detection \cite{tagarelli2017ensemble, zhuang2019dynamo}. 
Most of the ML/DL applications are underlying the assumption that the training and testing data are drawn from the same distribution (domain). However, in practice, it is more common that the data are from various domains. For instance, the image data for the medical diagnosis application might be collected from different hospitals, by different types of devices, or using different data preprocessing protocols. The domain shift issue results in a rapid performance degradation, where the machine learning applications is trained on ``seen'' domains and tested on other ``unseen'' domains. Even well-known strong learners such as deep neural networks are sensitive to domain shifts \cite{dann}.  It is crucial to enhance the generalization capability of machine learning models in the real-world applications. Because, on one hand, it is costly to re-collect/label the data and re-train the model for such ``unseen'' domains.  On the other hand, we can never enumerate all the ``unseen'' domains in advances. 

\begin{figure}[t]
  \centering
  \includegraphics[width=\linewidth]{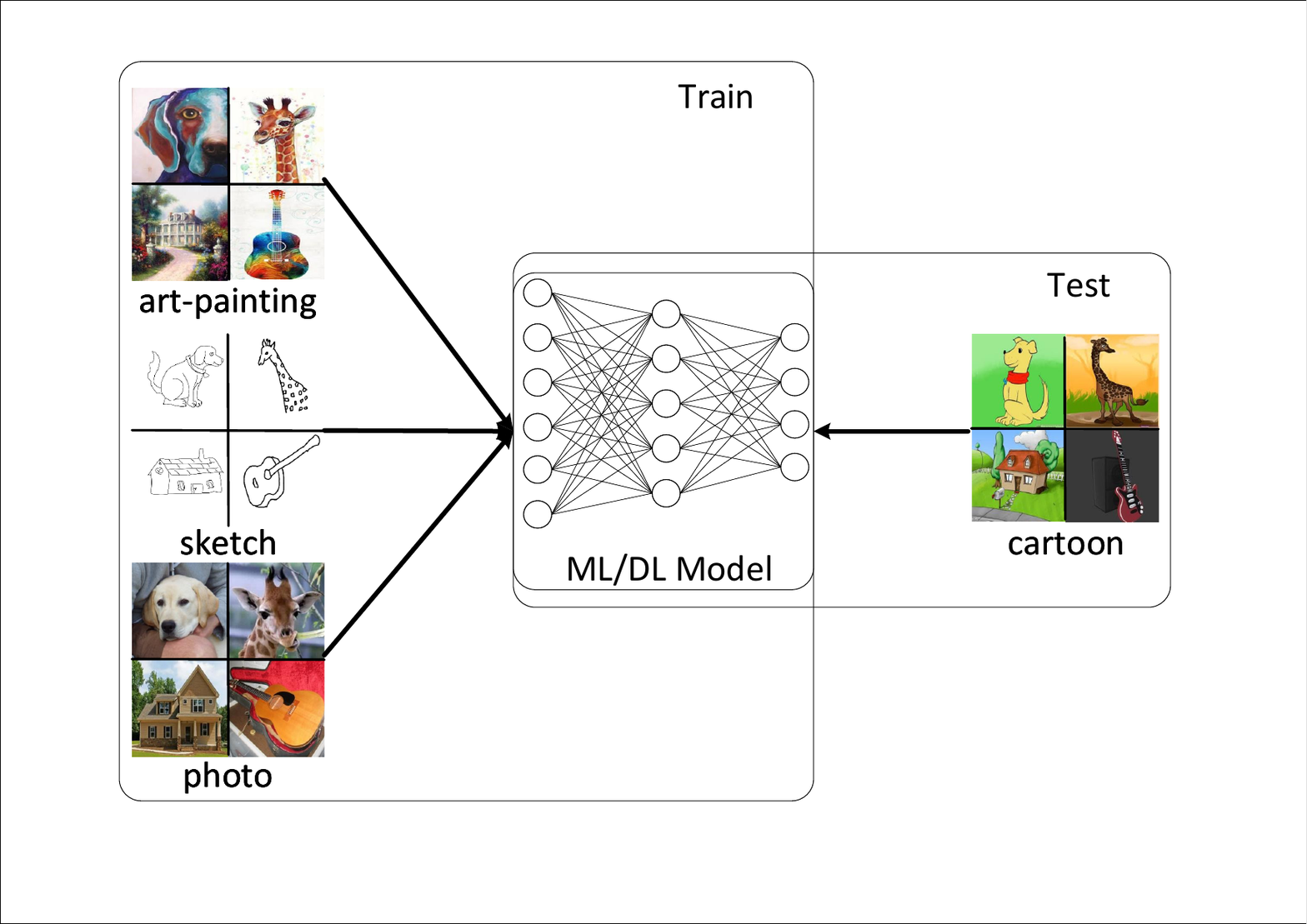}
  \caption{Multi-source Domain Generalization: training a model on one or multiple seen source domains and test on certain ``unseen'' target domain.}
  \label{DGdiagram}
\end{figure}

Domain Generalization (DG), as illustrated in Figure~\ref{DGdiagram}, which aims to learn a domain-invariant feature representation from multiple given domains and expecting good performance on the “unseen” domains.
It is one of the techniques that aiming to enhance the generalization capability of machine learning models. 
However, designing an effective domain generalization approach is challenging.
First, a well-designed DG approach should be model-agnostic. Domain shift is a general problem in the designing of ML/DL models, such that the approach should not be designed for a specific network architecture.
Second, an effective DG approach should not be data-dependent. There exists different types of domain shift, such as different art forms or different centric-images. A data-dependent approach can lead promising results on some datasets. However, the approach can be overfitting to the particular domain shift and might not have comparable  performance on the other datasets.
Hence, it is a challenging task to design an effective DG approach.

To date, a few algorithms have been proposed to enhance the generalization capability of ML/DL models. For instance,
D-SAM \cite{dsam} designs a domain-specific aggregation module for each ``seen'' domain, and plugs it on a particular network architecture to eliminate the domain specific information. However, it is a model-based approach, because the aggregation module is designed for a particular model, and additional implementation of aggregation module is required when the model changed.
Hex \cite{hex} is proposed to learn robust representations cross various domains via reducing the model dependence on high-frequency textural information. The original supervised model is trained with an explicit objective to ignore the superficial statistics, which only presents in certain datasets. Its representation learning is fully unsupervised, and performs good on certain image datasets. However, due to the assumption of domain shift and the unsupervised natural, Hex might not have the promising performance on the other image datasets. 
Approaches that leveraging the idea of meta-learning for domain generalization have been also proposed \cite{mldg, feature-critic, metareg}. For instance, MLDG \cite{mldg} was inspired by MAML \cite{maml} to simulate the domain-shift and optimize meta-train and meta-test together during the training phase. However, it only focuses on the classifier optimization, and lacks of effective guidance on the feature representation learning, where the better feature representation can benefit the classifier to make decisions. 

In this paper, we present a novel DG approach, Discriminative Adversarial Domain Generalization (DADG). 
Our DADG contains two main components, discriminative adversarial learning (DAL) and meta-learning based cross domain validation (Meta-CDV). 
We adopt the DAL to learn the set of features, which provides domain-invariant representation for the following classification task, and apply the Meta-CDV to further enhance the robustness of the classifier.
Specifically, on one hand, we consider the DAL conponent as a discriminator that trains a domain-invariant feature extractor by distinguishing the source domain of corresponding training data. 
On the other hand, we employ meta-learning optimization strategy to ``boost'' the objective task classifier by validating it on previously ``unseen'' domain data in each iteration.
The two components guide each other from both feature representation and object recognition level via a model-agnostic process over iterations to build a domain-generalization model.
Note that our DADG makes no assumption on the datasets, and it is a model-agnostic approach, which can be applied to any network architectures.

In the experimental evaluation, a comprehensive comparison has been made among our DADG and other 8 existing DG algorithms, including DeepAll (i.e., the baseline that simply used pre-trained network, without applying any DG techniques), TF \cite{tf}, Hex \cite{hex}, D-SAM \cite{dsam}, MMD-AAE \cite{mmd-aae}, MLDG \cite{mldg}, Feature-Critic (FC) \cite{feature-critic}, and JiGen \cite{jigsaw}. We conduct the comparison and the evaluation of our approach on three well-known DG benchmark datasets: PACS \cite{tf}, VLCS \cite{VLCS} and Office-Home \cite{office-home}, utilizing two deep neural network architectures, AlexNet and ResNet-18. Our experimental result shows that our approach performs well at cross domain recognition tasks. Specifically, we achieve the best performance on 2 datasets (VLCS and Office-Home) and performs $2^{nd}$ best on PACS. For instance, on VLCS dataset, we improve on the strong baseline DeepAll by 2.6\% (AlexNet) and 3.11\% (ResNet-18). Moreover, an ablation study also conducted to evaluate the influence of each component in DADG.

To summarize, our work has the following contributions:

$\bullet$ We present a novel, effective and model-agnostic framework, Discriminative Adversarial Domain Generalization (DADG) to tackle the DG problem. Our approach adopts discriminative adversarial learning to learn the domain-invariant feature extractor and utilizes meta-learning optimization strategy to enhance the robustness of the classifier.

$\bullet$ To the best of our knowledge, DADG is the first work that uses meta-learning optimization to regularize the feature learning of discriminative adversarial learning in domain generalization. 

$\bullet$ A comprehensive comparison among our algorithm and the state-of-the-art algorithms has been conducted (Section~\ref{section:exp}). For the sake of reproducibility and convenience of future studies about domain generalization, we have released our prototype implementation of DADG. \footnote[1]{\url{https://github.com/keyu07/DADG}}

The rest of this paper is organized as follows:
Section~\ref{section:related_work} presents the related literature review. 
Section~\ref{section:method} presents the notations in common domain generalization problem, and describes our proposed algorithm. 
Section~\ref{section:exp} presents the experimental evaluation. 
Section~\ref{section:conclusion} presents the conclusion.

\section{Related Work}\label{section:related_work}

\subsection{Generative Adversarial Nets (GAN)} \label{sub:GAN}
Generative Adversarial Nets (GAN) \cite{gan} aims to approximate the distribution $P_{d}$ of a dataset via a generative model. GAN simultaneously trains two components generator $G$ and discriminator $D$. The two components, generator and discriminator can be built from neural networks (e.g., convolutional layers and fully connected layers). The input of $G$ is sampled from a prior distribution $P_{z}(z)$ through which $G$ generates fake samples similar to the real samples. Meanwhile, $D$ is trained to differentiate between fake samples and real samples, and sends feedback to $G$ for improvement. GAN can be formed as a two-player minimax game with value function $V(G,D)$:

\begin{equation}
    \begin{split}
    \underset{G}{min} \ \underset{D}{max} \ V(G,D) = E_{x \sim P_{d}}[log(D(x))]+\\
    E_{z \sim P_{z}}[log(1-D(G(z)))]
    \end{split}
\end{equation}

GAN-based discriminative adversarial learning is able to learn a latent space from multiple different domains, where the latent space is similar to the given domains. It has been used in some domain adaptation works, which we will discuss below.

\subsection{Domain Adaptation} \label{sub:DA}
Domain adaptation (DA) is one of the closely related work to domain generalization. The main difference between DA and DG is that DA assumes unlabeled target data is available during the training phase, but DG has no access to the target data. Many domain adaptation algorithms \cite{adda, dann, unsupervised-DA, sim-domain} are designed via mapping the source and the target domain into a domain-invariant feature space. GAN or GAN-based discriminative adversarial techniques have been utilized in many such domain adaptation works. For instance, ADDA \cite{adda} maps the data from the target domain to the source domain through training a domain discriminator. DANN \cite{dann} is proposed to train a ``domain classifier'' to learn the latent representations of the source and the target domains. Tzeng et al. \cite{sim-domain} proposes to use multiple adversarial discriminators to apply on the data of different available source domains. 
Discriminative adaversarial learning successfully learns the domain-invariant feature representation, which considered as a latent space that similar to all source domains. 
This success motivates us to optimize the feature learning of domain generalization.

\subsection{Domain Generalization} \label{sub:DG}
In contrast to domain adaptation, domain generalization is a more challenging problem, because it requires no prior knowledge about the target domain.
Given a ML/DL application that has multiple ``seen'' or/and ``unseen'' domains, we observe that each domain has two elements: the private element and the global element. The private element contains the specific representation/information of each domain, while the global element holds the invariant features across different domains. 
Most of the recent domain generalization works aim to improve the learnt feature by using one of the two strategies: (i) Eliminating the influence of the private elements or (ii) Extracting the global elements. 
Other than the two main strategies, there are other alternative studies, such as a data augmentation based method \cite{domain-randomization} and a recent self-supervised learning method JiGen \cite{jigsaw}. JiGen \cite{jigsaw} uses a jigsaw-puzzle classifier to guide the feature extractor to capture the most informative part of the images, and it achieves current state-of-the-art results on three domain generalization benchmark datasets. We include JiGen \cite{jigsaw} in all our evaluations.

Many model-enhancement based studies are proposed under the first strategy. 
For instance, 
Li et al. \cite{tf} develops a low-rank parameterized network to decrease the size of parameters.
D’Innocente et al. \cite{dsam} proposes to build domain-specific aggregation modules and stack on the backbone network to merge specific and generic information. However, it is a model based approach. Because one set of aggregation modules can only apply on one particular backbone network. Additional implementation is required when we change the network architecture.
Hex \cite{hex} is proposed to learn robust representations cross various domains via reducing the model dependence on high-frequency textural information. The original supervised model is trained with an explicit objective to ignore the so called superficial statistics, which is presented in the training set but may not be present in future testing sets. Its representation learning is fully unsupervised, and performs good on certain image datasets. However, because the assumption of domain shift and the unsupervised natural of Hex, it might not have the comparable good performance on the other image datasets. 
However, designing an approach to weaken certain types of domain-specific elements may suffer from overfitting on such domain elements. 
Though some outstanding results have been shown by this kind of approaches on certain datasets, while may not be able to be generalized to many more ``unseen'' domains. For instance, the different domain types are considered as different art forms or different centric-images.

For the second strategy, most of the previous works are focusing on learning domain-invariant representation, which is able to capture the important similar information among multiple different domains and have the capability of generalizing to more ``unseen'' domains. 
As such, these works are more similar to the work of domain adaptation. 
For intance, Ghifary et al. \cite{mtae} proposes to learn domain-invariant features via a multi-domain reconstruction auto-encoder. However, the effectiveness for reconstruction the auto-encoder is limited while applying to more complex datasets \cite{tf}. 
Motiian et al.\cite{ccsa} employs maximum mean discrepancy (MMD) and proposes to learn a latent space that minimizes the distance among images that have the same class label but different domains.
Li et al. \cite{mmd-aae} proposes to align source domains to learn a domain-agnostic representation using adversarial autoencoders with MMD constraints, and uses adversarial learning to match the distribution of generated data with a prior distribution.

Our approach also belongs to the second strategy. We use the discriminative adversarial learning to learn a latent distribution among the source domains. By doing so, we achieve a domain-invariant feature representation that different domains are indistinguishable. Beyond the domain-invariant feature representation, in order to improve the relevant classification task, we also propose a more robust classifier, by using meta-learning based optimization, which leads more competitive classification results. To the best of our knowledge, this is the first work that uses meta-learning optimization to regularize the discriminative adversarial learning in domain generalization.

\subsection{Meta-Learning} \label{sub:meta-learning}
Meta-learning introduces a concept ``learning-to-learn'' and recently receives great interests with applications including few-shot learning \cite{maml,reptile,meta-implicit} and learning optimizations \cite{li2017learning,gdbygd}. It learns from various tasks during training and such that the model can be quickly generalized to new tasks. 
MAML \cite{maml} is typical in those works. It utilizes sampled episodes during training, where each episode is designed to simulate the few-shot tasks in a train-test split manner. 
Recently, a few works have applied this episodic meta-learning optimization method in domain generalization \cite{mldg,metareg,feature-critic}. For instance, MLDG \cite{mldg} borrows the idea of \cite{maml} to optimize the classifier, by simulating the train-test domain shift during training phase.
MetaReg \cite{metareg} proposes to learn a regularization function for the network classifier.
Li et al. \cite{feature-critic} proposes to simultaneously learn an auxiliary loss and measure whether the performance of validation set has been improved. 
However, MLDG \cite{mldg} and MetaReg \cite{metareg} only focus on classifier optimization, and are lacking of details addressing the learning of a domain-invariant feature space.
The success of meta-learning method on the enhancement of classifier robustness motivates us to optimize the network classifier for domain generalization.
To summarize, in order to address the challenging domain generalization problem, we apply discriminative adversarial learning and meta-learning, where the discriminative adversarial learning extracts domain-invariant feature representation, and meta-learning enhances the classifier robustness.


\section{Methodology} \label{section:method}
The design of DADG is based on our assumption that there exists a domain-invariant feature representation, which contains the common  information for both the ``seen'' and ``unseen'' domains. It should satisfy the following properties: 
(i) The feature representation should be invariant in terms of data distributions (domains). Since ML/DL models are designed to transfer the knowledge from seen domains to unseen domains, they could fail if the distributions differ a lot.
(ii) It should keep the variance between different objects (classes). This helps the model to capture the unique information of different objects and to make precise decisions. We use two key components in DADG to address the above two properties: discriminative adversarial learning (DAL) and meta-learning based cross domain validation (Meta-CDV).
DAL aims to learn a domain-invariant feature representation where different data distributions are indistinguishable. Therefore, the domain variance will be minimized. 
Meta-CDV brings the learnt features to supervised learning by training a classifier in a meta-learning manner. It evaluates the validation performance of previous unseen domains within each training iteration. 

\begin{figure}[t]
  \centering
  \includegraphics[width=\linewidth]{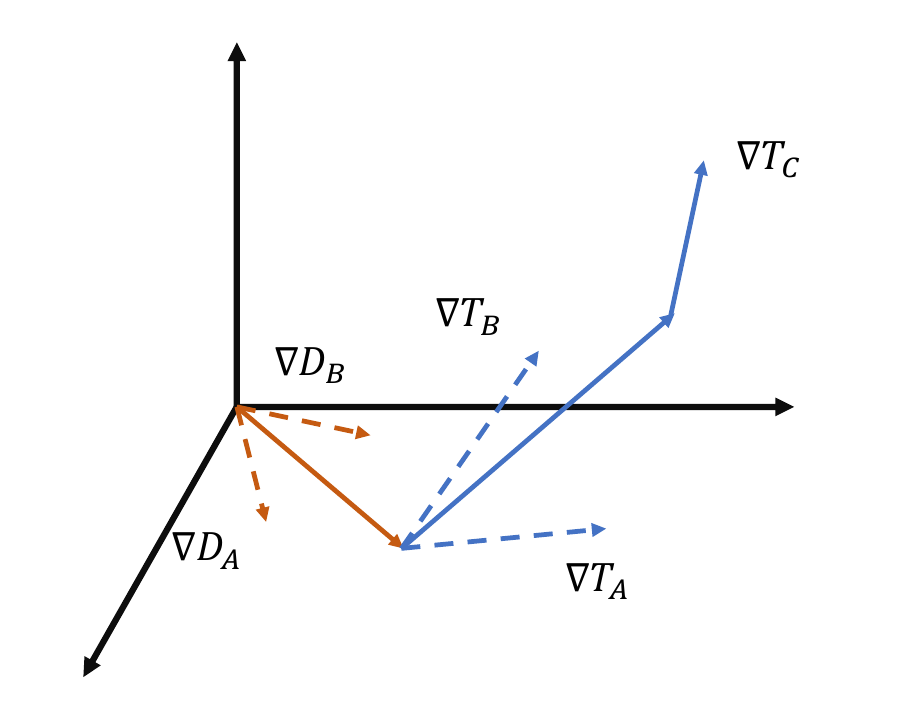}
  \caption{Diagram of our DADG. Better view in colors.}

  \label{3dupdate}
\end{figure}

We introduce Figure \ref{3dupdate} to better illustrate our DADG in high level. The goal of DADG is to find the optimized feature representation point, which satisfies the two properties. $A$, $B$ and $C$ present the different domains. DAL and Meta-CDV address DG in two aspects: 
(i) As shown by the orange lines, the dash lines are the gradient directions when tackling feature learning on different domains $\nabla D_A$ and $\nabla D_B$, respectively. While the solid line is the actual gradient direction guided by DAL and finally reaches a representation point indistinguishable from given domains.
(ii) As shown by the blue lines, the dash lines indicate the gradient directions when solving certain tasks $\nabla T_A$ and $\nabla T_B$, respectively. While the solid line denotes the actual gradient direction led by classification task on two domains and further optimized by cross domain validation ($\nabla T_C$). 
The model finally learns a domain-invariant feature representation point that satisfies the two properties.

In the rest of this section, we denote the input data space as $x \in X$, the class label space as $y \in Y$ and the domain label (i.e., belonging to which distribution) space as $y^d \in Y^d$. The source domains are described as 
$D_i \in S$, and the target domains as $T$. Also, please note that in the rest of this section, the superscript of each parameter indicates different updating stages within one iteration, denoted as $m$, while the subscript indicates different iterations, denoted as $n$. We introduce our two main components in the remaining sections:  DAL in \ref{DAL} and Meta-CDV in \ref{Meta-CDV}. Finally we summarize the two components together in \ref{approachsummary}. 

\subsection{Discriminative Adversarial Learning} \label{DAL}
As described above, the goal of this component is to learn a domain classification model, which aims to classify data from different domains. We consider our DAL containing two parts: (i) a feature extractor \textit{$f_{\theta}$} with parameter \textit{$\theta$}, and (ii) a discriminator \textit{$d_{\psi}$} with parameter $\psi$. Both $\theta$ and $\psi$ are learnable parameters during training phase.

In our approach, we first randomly divide the source domains $S$ into two mutually exclusive sets: 
$S_d$ for DAL and $S_c$ for Meta-CDV.
The discriminator acts as a domain classifier, which takes the learnt sample features $f_\theta(x_j)$ of each arbitrary input $x_j$ and tries to discriminate its domain label $y^d$. Thus, we need to learn the parameters ($\psi$) that minimize the classification loss, which as follows:
\begin{equation}
   \textit{$\mathcal{L}_{disc}(d_{\psi^{m}_n}(f_{\theta^{m}_n}(x_j)), y^d_j)$}
\end{equation}

The loss function of DAL is presented as follows:

\begin{equation}
   \textit{$F(\cdot) = \sum_{D_i\in S_d}\sum_{x_j\in D_i} \mathcal{L}_{disc}(d_{\psi^{m}_n}(f_{\theta^{m}_n}(x_j)), y^d_j)$}  \label{object_d}
\end{equation}

The objective of the feature extractor is to maximize the discriminative loss, to achieve indistinguishable of the learnt feature representation. 
Following the design of GAN \cite{gan}, the objective function of our discriminative adversarial learning can be written as the following minimax optimization:
\begin{equation}
   \textit{$\underset{\psi^{m}_n}{argmin}~\underset{\theta^{m}_n}{max}~F(\cdot)$}  \label{minmax}
\end{equation}

Such minimax parameter updating can be achieved by gradient reversal layer (GRL) \cite{unsupervised-DA}, which placed between the feature extractor and discriminator. During forward propagation, GRL keeps the learnable parameters same. During back propagation, it multiply the gradient by $-\lambda$ and pass it to the preceding layer. 

To summarize, we update the parameters of feature extractor and discriminator as follows:
\begin{equation}
   \textit{$\theta^{m+1}_n \gets \theta^{m}_n - \alpha \cdot \nabla(-\lambda \cdot F(\cdot)) $}  \label{domain_e}
\end{equation}
\begin{equation}
   \textit{$\psi_{n+1}^{m} \gets \psi^{m}_n - \alpha \cdot \nabla F(\cdot) $}  \label{domain_d}
\end{equation}
where the $\alpha$ is the DAL learning rate. Thereafter, the $\theta^{m+1}_n$ will be shared in further training within the same iteration (as we illustrated in Figure \ref{flow} step \textcircled{1}), and $\varphi^m_{n+1}$ will be used in the next iteration.

\subsection{Meta-learning based Cross Domain Validation} \label{Meta-CDV}
After the feature extractor has been trained to minimize the domain variance, we adopt meta-learning based cross domain validation (Meta-CDV) to address the enhancement of the classifier robustness. Robust classifier is able to help the feature extractor to keep the discriminant power between various classes. This is accomplished by training the classification model on 2 seen domains $S_d$ in DAL and validating the performance on cross domains $S_c$. 


To train the model on seen domains $S_d$, the classification model is composed of the feature extractor \textit{$f_{\theta}$} from DAL and a classifier \textit{$c_{\varphi}$} with parameters $\varphi$. The training loss is defined as follows:
\begin{equation}
   \textit{$\mathcal{L}_{train}(c_{\varphi^{m}_n}(f_{\theta^{m+1}_n}(x_j), y_j)$}
\end{equation}

where $x_j$ is an arbitrary input and $y_j$ is the corresponding output label. 

\begin{figure}[t]
  \centering
  \includegraphics[width=\linewidth]{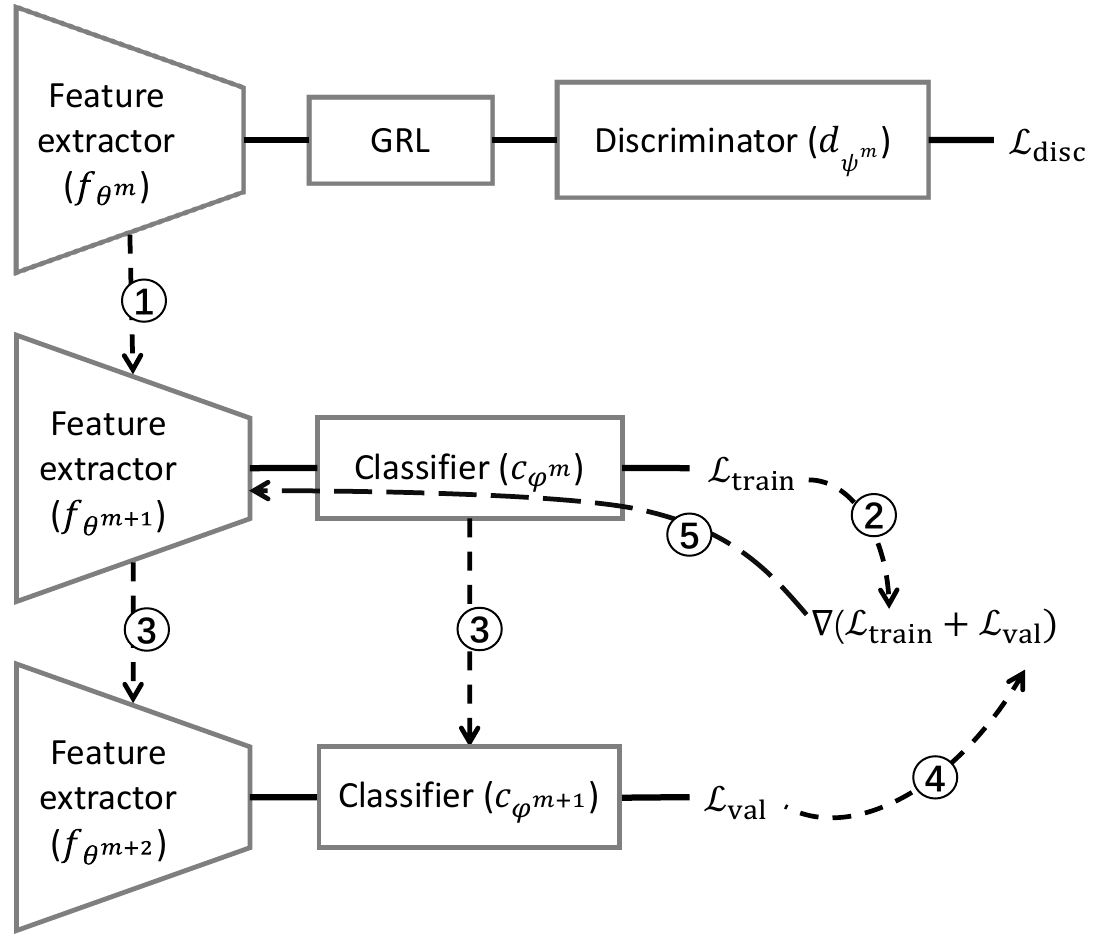}
  \caption{The training flow of DADG.}
  \label{flow}
\end{figure}

The loss function of classification training on seen domains is presented as follows (as illustrated in Figure \ref{flow} step \textcircled{2}):
\begin{equation}
   \textit{$G(\cdot) = \sum_{D_i\in S_d}\sum_{x_j\in D_i} \mathcal{L}_{train}(c_{\varphi^{m}_n}(f_{\theta^{m+1}_n}(x_j)), y_j)$}  \label{train_ob}
\end{equation}

Note that the training is performed over the updated feature extractor parameters $\theta^{m+1}$ in DAL. As such, the parameters are updated as follows:
\begin{equation}
   \textit{$\theta^{m+2}_n \gets \theta^{m+1}_n - \beta \cdot \nabla~G(\cdot) $}  \label{train_e}
\end{equation}
\begin{equation}
   \textit{$\varphi^{m+1}_n \gets \varphi_n^m - \beta \cdot \nabla~G(\cdot) $}  \label{train_c}
\end{equation}
where the $\beta$ is the classification learning rate. Here the updated parameter $\theta^{m+1}_n$ is involved in the calculation of training loss. It also means that we need the second derivative with respect to $\theta$, while minimizing the loss function \ref{train_ob}.

After finishing the classification task on seen domains, we evaluate the performance on cross domains $S_c$ to boost the classification model. This process simulates the virtual train/test settings. The evaluation is performed on the updated parameters $\theta^{m+2}_n$ and $\varphi^{m+1}_n$ (as illustrated in Figure \ref{flow} step \textcircled{3}). More concretely, this evaluation come up with the cross domain validation loss: 
\begin{equation}
   \textit{$\mathcal{L}_{val}(c_{\varphi^{m+1}_n}(f_{\theta^{m+2}_n}(x_j)), y_j)$}
\end{equation}

The loss function of cross domain validation is as follows:
\begin{equation}
   \textit{$H(\cdot) = \sum_{D_i\in S_c}\sum_{x_j\in D_i} \mathcal{L}_{val}(c_{\varphi^{m+1}_n}(f_{\theta^{m+2}_n}(x_j)), y_j)$}  \label{val}
\end{equation}

Finally, as illustrated in Figure \ref{flow} step \textcircled{2}, \textcircled{4} and \textcircled{5}, we update our classification model by adding the training loss $\mathcal{L}_{train}$ and cross domain validation loss $\mathcal{L}_{val}$ at the end of each iteration:
\begin{equation}
   \textit{$\theta_{n+1}^{m} \gets \theta^{m+1}_{n} - \gamma \cdot \nabla H(\cdot) $}  \label{total_e}
\end{equation}
\begin{equation}
   \textit{$\varphi_{n+1}^{m} \gets \varphi^{m}_{n} - \gamma \cdot \nabla H(\cdot)$}  \label{total_c}
\end{equation}

where $\gamma$ presents the learning rate of cross domain validation.
Note that the parameter updating on seen domains classification is performed over the parameter $\theta^{m+1}_n$ and $\varphi^m_n$, whereas the cross domain validation is evaluated over parameter $\theta^{m+2}_n$ and $\varphi^{m+1}_n$. In other words, the optimization of our classification model is involved in third derivative with respect to $\theta$ and second derivative with respect to $\varphi$. 

\subsection{Summary of DADG} \label{approachsummary}
As illustrated in Figure~\ref{flow}, the DAL and Meta-CDV optimize the model by addressing different aspects of domain generalization, and work synergistically within one iteration. In each iteration, we randomly split the train/validation ($S_d$/$S_c$) domains. DAL learns a domain-invariant feature extractor (\textit{$f_{\theta}$}) by maximizing the discriminative loss. Then, our approach learns a robust classification model by adopting a simple classification training and cross domain validation, which optimized in meta-learning based manner.
For the whole process, the objective function can be introduced as: 
\begin{equation}
   \textit{$\underset{\psi^{m}_n}{argmin}~\underset{\theta^{m}_n}{max}~F(\cdot) + \underset{\theta^{m+1}_n, ~\varphi^m_n}{argmin}~(G(\cdot) + H(\cdot))$}  \label{object_final}
\end{equation}

Once Equation \ref{object_final} is optimized to converge on the source domains, we evaluate the classification model using unseen domains.

\section{Experimental Evaluation} \label{section:exp}
We conduct our experiments on 3 benchmark datasets (PACS \cite{tf}, VLCS \cite{VLCS} and Office-Home \cite{office-home}) and 2 deep neural network architectures with pretrained parameters (AlexNet \cite{krizhevsky2012imagenet} and ResNet-18 \cite{he2016deep}) to evaluate the generalization capability of our proposed approach. A comprehensive comparison has been made among our approach and other baseline approaches.
The presented results are shown that our DADG performs consistently comparable in all the evaluations, and achieves the state-of-the-art results in two datasets. The effectiveness of each component in our approach also discussed. All the details are described in following.

\subsection{Baseline Approaches} \label{sub:competitor}
We compare our proposed approach performance with following baseline DG approaches. 
\begin{itemize}
    \item \textbf{DeepAll} is the baseline that simply use deep learning network to train the aggregation of all source domains and test the unseen domain. It is a strong baseline that surpasses many previous DG works \cite{tf}. 
    \item \textbf{TF} \cite{tf} introduces a low-rank parameter network to decrease the size of parameters. This work also shows that the DeepAll can surpass many previous studies and first provides PACS dataset.
    \item \textbf{Hex} \cite{hex} attempts to reduce the sensitivity of a model on high frequency texture information, and thus to increase model domain-robustness. 
    \item \textbf{MMD-AAE} \cite{mmd-aae} is based on adversarial autoencoder. It aligns different domain distributions to an arbitrary prior via MMD regularization, to learn an invariant feature representation.
    \item \textbf{Feature-Critic(FC)} \cite{feature-critic} aims to train a robust feature extractor. It uses meta-learning approach, along with an auxiliary loss to measure whether the updated parameter has improved the performance on the validation set. 
    \item \textbf{MLDG} \cite{mldg} is the first work that addresses domain generalization using meta-learning. It is inspired by MAML \cite{maml} and proposed visual cross domain classification task by splitting source domains into meta-train and meta-test. 
    \item \textbf{D-SAM} \cite{dsam} plugs parallel domain-specific aggregation modules on a given network architecture to neglect domain specific information.  
    \item \textbf{JiGen} \cite{jigsaw} is the first work that addresses DG by self-supervised learning. It divides each image into small patches and shuffle the order. Then, trains an object classifier and a jigsaw order classifier simultaneously. It achieves the state-of-the-art results on the three datasets VLCS \cite{VLCS}, PACS \cite{tf} and Office-Home \cite{office-home}. 
\end{itemize}

\subsection{Experimental Datasets} \label{sub:dataset}
We utilize three well-known domain generalization benchmark datasets. 
\begin{itemize}
 \item \textit{\textbf{VLCS}} \cite{VLCS} is composed of 10,729 images with resolution 227 $\times$ 227, taken from 4 different datasets (i.e., domains): PASCAL \textbf{V}OC2007 \cite{voc}, \textbf{L}abelMe \cite{labelme}, \textbf{C}altech101 \cite{caltech} and \textbf{S}un09 \cite{sun}. It depicts 5 categories (i.e., classes): bird, car, chair, dog and person. 
 \item \textit{\textbf{PACS}} \cite{tf} contains more severe domain shifts than VLCS. PACS aggregates 9,991 images in 7 different classes: dog, elephant, giraffe, guitar, house, horse and person. It shared by 4 different domains: \textbf{P}hoto, \textbf{A}rt, \textbf{C}artoon and \textbf{S}ketch. 
 \item \textit{\textbf{Office-Home}} \cite{office-home} was created to evaluate DA and DG algorithms for object recognition in deep learning. There are 15,592 images from 4 different domains: Art, Clipart, Product and real-world images, each domain includes 65 classes.
\end{itemize}

\subsection{Experimental Setting} \label{sub:exp_setting}
All three benchmark datasets contain the data of four different domains. We first hold one domain (i.e., the target domain) for testing and the rest three for training. Then, in the training phase, we randomly select two domains to apply discriminative adversarial learning (DAL), and select one domain to boost our classifier by meta-learning based cross domain validation (Meta-CDV). Our discriminator consists of two fully connected layer with 1024 neurons each and one output layer with 1 neuron.

The neural network is updated by stochastic gradient descent(SGD) in 2000 iterations during training. We use cross-entropy loss for both DAL (domain classification task) and Meta-CDV (classification task). Negative log-likelihood loss also tested for classification task, but it hardly effects the performance. 

For most the hyperparameters, we followed MLDG: base classification learning rate $\beta = 5 \times 10^{-4}$, cross domain validation learning rate $\gamma = 5 \times 10^{-4}$, momentum = 0.9 and weight decay = $5 \times 10^{-5}$. The DAL learning rate is $\alpha = 5 \times 10^{-5}$. While a big $\alpha$ value will lead an unstable training process and 5 $\times 10^{-5}$ is appropriate for PACS, VLCS and Office-Home. The value of $\alpha$ should be picked carefully on the other datasets, 1/10 of the $\beta$ and $\gamma$ is suggested. The model-agnostic can be achieved by simply changing the backbone network architectures without additional implementation. All of our experiments are implemented using PyTorch, on a server with GTX 1080Ti 11 GB GPU.

\begin{table}
    \caption{Cross domain classification accuracy (in \%) on VLCS dataset when using network architecture AlexNet and ResNet-18. The results of our implementation were the average over 20 repetitions. Each column name indicates the target domain. Best performance in bold.}
    \label{tab1}
    \scalebox{0.82}{
        \begin{tabular}{cccccc}
            \toprule
            \textbf{VLCS} & \textbf{VOC} & \textbf{LabelMe} & \textbf{Caltech} & \textbf{Sun}&\textbf{Avg.}\\
            \midrule
            \multicolumn{6}{c}{\textbf{AlexNet}} \\
            \midrule
            {TF} \cite{tf}  & 69.99 & 63.49  & 93.63 & 61.32 & 72.11\\
            HEX$^*$ \cite{hex}  & 68.51  & \textbf{63.67}  & 89.63  & 62.12 & 70.98 \\
            MMD-AAE \cite{mmd-aae}  & 67.70 & 62.60 & 94.40 & 64.40 & 72.28\\
            FC$^*$ \cite{feature-critic}  & 66.79  & 61.48  & 95.68  & 63.13 & 71.77\\
            MLDG$^*$ \cite{mldg}  & 70.01  & 61.06  & 95.68  & 65.08 & 72.96 \\
            {D-SAM} \cite{dsam}  & 63.75 & 54.81  & 94.96 & 64.56 & 69.52\\
            {JiGen} \cite{jigsaw}  & 70.62 & 60.90 & \textbf{96.93} & 64.30 & 73.19 \\
            \midrule
            DeepAll  & 68.11  & 61.30  & 94.44  & 63.58 & 71.86\\
            DADG  & \textbf{70.77} & 63.44  & 96.80 & \textbf{66.81} & \textbf{74.46} \\
            \midrule
            \multicolumn{6}{c}{\textbf{ResNet-18}} \\
            \midrule
            MLDG$^*$ \cite{mldg}  & 74.41  & 63.45  & 96.75  & 69.35 & 75.99\\
            D-SAM$^*$ \cite{dsam} & 70.42  & 58.70  & 88.90  & \textbf{71.36} & 72.35\\
            JiGen$^*$ \cite{jigsaw}  & 74.91  & 63.00  & 98.39  & 69.37 & 76.42\\
            \midrule
            DeepAll  & 73.84 & 62.17 & 97.10  & 67.28 & 75.10\\
            DADG  & \textbf{76.17} & \textbf{67.22} & \textbf{98.50} & 70.95 & \textbf{78.21} \\
            \bottomrule
        \end{tabular}}
\end{table}

\begin{table}
  \renewcommand\arraystretch{1.1}
  \caption{Cross domain classification accuracy (in $\%$) on PACS dataset when using network architecture AlexNet and ResNet-18. The results of our implementation were the average over 20 repetitions. Each column name indicates the target domain. Best performance in bold.}
  \label{tab2}
  \scalebox{0.83}{
  \begin{tabular}{cccccc}
    \toprule
    \textbf{PACS} & \textbf{Photo} & \textbf{Art-paint} & \textbf{Cartoon} & \textbf{Sketch}&\textbf{Avg.}\\
    \midrule
    \multicolumn{6}{c}{\textbf{AlexNet}} \\
    \midrule
     {TF} \cite{tf}  & 89.50 & 62.86  & 66.97 & 57.51 & 69.21\\
     {HEX} \cite{hex}  & 87.90 & 66.80 & 69.70 & 56.30 & 70.18\\
     {FC} \cite{feature-critic}  & \textbf{90.10} & 64.40  & 68.60 & 58.40 & 70.38\\
     {MLDG}\cite{mldg}  & 88.00 & 66.23 & 66.88 & 58.96 & 70.02\\
     {D-SAM} \cite{dsam}  & 85.55 & 63.87  & 70.70 & 64.66 & 71.20 \\
     {JiGen} \cite{jigsaw}  & 89.00 & \textbf{67.63} & \textbf{71.71} & \textbf{65.18} & \textbf{73.38}\\
    \midrule
     DeepAll  & 88.65  & 63.12  & 66.16  & 60.27 & 69.55\\
     DADG  & 89.76 & 66.21 & 70.28 & 62.18 & 72.11 \\
    \midrule
    \multicolumn{6}{c}{\textbf{ResNet-18}} \\
    \midrule
     MLDG$^*$ \cite{mldg}  & 94.03  & 76.42  & 73.03 & 68.15 & 77.91\\
     D-SAM \cite{dsam} & 95.30  & 77.33  & 72.43 & \textbf{77.83} & \textbf{80.72}\\
     {JiGen} \cite{jigsaw}  & \textbf{96.03}  & 79.42  & 75.25  & 71.35 & 80.51\\
    \midrule
     DeepAll  & 93.06 & 75.60  & 72.30  & 68.10 & 77.27\\
     DADG  & 94.86 & \textbf{79.89}  & \textbf{76.25} & 70.51 & 80.38 \\
    \bottomrule
\end{tabular}}
\end{table}

\subsection{Effectiveness Analysis} \label{sub:results}
In this section, we discuss the performance of our proposed approach and the baseline approaches in terms of classification accuracy. Table \ref{tab1} - Table \ref{tab3} show the results of datasets VLCS, PACS and Office-Home. To make a more comprehensive comparison, we implement MLDG our own, because only demo code is provided by the author. Besides, we implement Hex, Feature-Critic, D-SAM and JiGen by using the code that are provided by the authors. All the implementations are evaluated on the datasets or network architectures they did not report. Our results of these approaches are highlighted in the three tables with *. The details of each dataset are presented below:

\textit{\textbf{VLCS:}} We follow the standard protocol of MTAE \cite{mtae} to randomly divide the data of each source domain into training (70\%) and testing (30\%) sets. Finally we test on all the images in target domain. The upper and bottom part of Table \ref{tab1} show the results when using different network architectures AlexNet and ResNet-18, respectively. 
From table \ref{tab1}, we can observe that 
(i) The baseline DeepAll performs competitively and surpasses many previous DG works on overall performance, such as HEX, Feature-Critic and D-SAM. But our approach outperforms DeepAll in all target domain cases and on different network architectures.
(ii) On AlexNet, our DADG performs better than DeepAll by 2.6\% and better than Jigen by 1.27\%, such that we achieve the new state-of-the-art result on VLCS dataset. More specifically, DADG provides the best results in two (i.e., VOC and SUN respectively) out of four target cases. 
(iii) On ResNet-18, DADG surpasses the previous SOTA result Jigen in average performance and performs the best in three out of four target domain cases.

\textit{\textbf{PACS}}: We follow the training protocol of TF, considering three domains as source domains and the remaining one as target. The evaluation results are shown in Table \ref{tab2}, we can see that: 
(i) On AlexNet, although we do not achieve the best performance on any target domain cases, our DADG provides consistently comparable results, and performs the $2^{nd}$ best in average results.
(ii) On ResNet-18, we have two best results on Art-paint (79.89\%) and Cartoon (76.25\%), and only slight worse (0.34\%) than the best JiGen in average performance.

\textit{\textbf{Office-Home:}} We follow the protocol of D-SAM, also considering three as source domains and the rest one as target. The results are shown in Table \ref{tab3}, and we can observe that: 
(i) The advantage of of D-SAM in average results originates from its results on Art and Clipart, but the rest two were lower than DeepAll. 
(ii) Our DADG achieves the best in two target cases and the best in average results, and improves the previous SOTA result Jigen by 1.02\%. 

\begin{table}
  \renewcommand\arraystretch{1.1}
  \caption{Cross domain classification accuracy (in \%) on Office-Home dataset when using ResNet-18. The results of our implementation were the average over 20 repetitions. Each column name indicates the target domain. Best performance in bold.}
  \label{tab3}
  \scalebox{0.8}{
  \begin{tabular}{cccccc}
    \toprule
    \textbf{Office-Home} & \textbf{Art} & \textbf{Clipart} & \textbf{Product} & \textbf{Real-World}&\textbf{Avg.}\\
    \midrule
    \multicolumn{6}{c}{\textbf{ResNet-18}} \\
    \midrule
     MLDG$^*$ \cite{mldg}  & 52.88  & 45.72  & 69.90  & 72.68  & 60.30\\
     D-SAM \cite{dsam} & \textbf{58.03} & 44.37 & 69.22 & 71.45 & 60.77\\
    JiGen \cite{jigsaw}  & 53.04 & 47.51 & \textbf{71.47} & 72.79 & 61.20\\
    \midrule
     DeepAll  & 54.31  & 41.41 & 70.31  & 73.03 & 59.77\\
     DADG  & 55.57 & \textbf{48.71} & 70.90 & \textbf{73.70} & \textbf{62.22}\\
    \bottomrule
\end{tabular}}
\end{table}
\noindent
\textit{\textbf{Summary of the Experimental Evaluation:}}  \label{sub:summary} From the experimental evaluation analyzed above, we conclude that: 
(i) DeepAll exceeds many previous approaches in different datasets. In general, only MLDG, JiGen and our DADG can outperform DeepAll in all three datasets. 
(ii) As we mentioned in Section~\ref{sub:DG}. The approaches that aim to neglect particular domain-specific information, may assist the model in some datasets but fail in others. For instance, HEX and D-SAM are better than DeepAll on PACS, but worse than DeepAll on VLCS. 
(iii) our DADG has consistently comparable results in all the datasets and achieves the SOTA results on VLCS and Office-Home, also the second best on PACS. On VLCS and Office-Home, DADG outperforms the previous SOTA JiGen all over 1\%.

\subsection{Impact of Different DADG Components} \label{section:impact_dif_comp}
In this section, we conduct an extended study using PACS dataset with network architecture AlexNet to investigate the impact of the two key components (i.e., DAL and Meta-CDV) in our proposed approach DADG. Specifically, we test the performance in terms of classification accuracy by excluding each component in our approach respectively. DADG-DAL only contained the discriminative adversarial learning (DAL) component and trained the classification model conventionally instead of in meta-learning manner. While DADG-CDV meant that we removed the DAL component and only updated the classification model parameters in meta-learning manner. 

From the results in Table \ref{tab4}, we can see that DADG-DAL and DADG-CDV consistently perform better than DeepAll, and our full version DADG surpasses both baseline models in average performance and in every target domain cases. In the comparison between DADG-DAL and DADG-CDV, the DADG-DAL consistently better than the DADG-CDV.
The results in Table \ref{tab4} show that: 
(i) Employing discriminative adversarial learning is able to effectively guide the feature extractor to learn the invariant features among multiple source domains.
(ii) Since the only difference between DeepAll and DADG-CDV is the updating manner. Thus applying meta-learning based cross domain validation can make the classification model more robust.
(iii) The full version DADG consistently performs the best in every single case, which has shown that combining domain invariant representation and robust classifier together helped the model to enhance generalization. 
(iv) The domain-invariant representation plays a more crucial role rather than the robust classifier. Because the invariant representation provides a easier task for the classifier to make decision. 

\begin{table}
  \renewcommand\arraystretch{1.1}
  \caption{Cross domain classification accuracy (in \%) on PACS dataset using AlexNet. The results of our implementation were the average over 20 repetitions. Each column name indicates the target domain. Best performance in bold.}
  \label{tab4}
  \scalebox{0.76}{
  \begin{tabular}{cccccc}
    \toprule
    \textbf{PACS} & \textbf{Photo} & \textbf{Art-paint} & \textbf{Cartoon} & \textbf{Sketch}&\textbf{Avg.}\\
    \midrule
    \multicolumn{6}{c}{\textbf{AlexNet}} \\
    \midrule
    DeepAll  & 88.65  & 63.12  & 66.16  & 60.27 & 69.55\\
    DADG-DAL  & 89.51  & 65.43  & 69.19 & 61.70  & 71.46\\
    DADG-CDV & 89.10 & 64.22 & 68.24 & 60.60 & 70.54\\
     DADG  & \textbf{89.76} & \textbf{66.21} & \textbf{70.28} & \textbf{62.18} & \textbf{72.11} \\
    \bottomrule
\end{tabular}}
\end{table}

\subsection{Impact of Linear Related Domains} \label{domain_relation_anay}
We assume that there exists a domain-invariant feature representation for both source and target domains. However, it is also possible that some target domains are less relevant or even irrelevant to the source domains. 

The domain types were considered as different art forms (art, cartoon in PACS) or different centric images (LabelMe and SUN in VLCS) in previous sections. It is very hard to define whether the target domain is less relevant to the source domains.
To explore this situation, we conduct an experiment using digit images in six different angles as six different domains. To be more specific, we adopt the MNIST \cite{lecun-mnisthandwrittendigit-2010} dataset and randomly chose 1,000 images in each class and trained with AlexNet \cite{alexnet}. We denote the digit images rotated with \ang{0} by $R_0$ and then rotate the digit images in a counter-clock wise direction by \ang{15}, \ang{30}, \ang{45}, \ang{60} and \ang{75}. 
Since the rotation angles are continues related, which means sometimes the target domains are out of the scope of source domains (irrelevant). For example, when the $R_0$ and $R_{15}$ are as the target domains, we consider that the target domains are out of the scope of source domains. 
During the training phase, 4 domains are selected as source domains and the rest 2 are target domains. For each iteration, our DADG randomly adopts 2 source domains for DAL and another 2 for Meta-CDV. The model performance will evaluated on the rest 2 target domains. A comparison is made among DeepAll, MLDG \cite{mldg} and DADG. 

From the results in Table \ref{tab5}, we can see that:
(i) For all 3 approaches, the performance are better when the target domains close to \ang{30} and \ang{45}, and much worse when the target domains close to \ang{0} and \ang{75}. 
(ii) Our DADG outperforms the other two approaches in 10 out of total 15 different cases and achieves the best overall average accuracy among the 3 approaches. 
The results show the performance drop when the target domains are irrelevant to the source domains. It happens to all the approaches and can be considered as a common situation in domain generalization.
Although our DADG outperforms other 2 in average, only 10/15 better than the MLDG. Compare to the performance on VLCS, PACS and Office-Home (Table \ref{tab1}, \ref{tab2}, and \ref{tab3}), our DADG does not show significant advantage on this experiment. 
Because we select 2 source domains to do discriminative adversarial learning (DAL), and the rest source domains will train with Meta-CDV. When the number of source domains increased, DAL only contributes small portion in each iteration. As we mentioned in the section \ref{section:impact_dif_comp} (iv), DAL plays a more critical role than Meta-CDV. Finally, if we have a great number of source domains, the contribution of DAL can be even negligible. Thus, our DADG is sensitive to the number of source domains.

\begin{table}[t]
  \renewcommand\arraystretch{1}
  \caption{Cross domain classification accuracy (in \%) on MNIST rotation dataset using AlexNet. The results of our implementation were the average over 10 repetitions. Best performance in bold.}
  \label{tab5}
  \scalebox{0.78}{
  \begin{tabular}{ccccc}
    \toprule
    \textbf{Source} & \textbf{Target} & \textbf{DeepAll} & \textbf{MLDG$^*$ \cite{mldg}} & \textbf{DADG} \\
    \midrule
    \multicolumn{5}{c}{\textbf{AlexNet}} \\
    \midrule
    $R_{30}, R_{45}, R_{60}, R_{75}$ & $R_{0}, R_{15}$  & 69.35 & 69.51  & \textbf{69.57}   \\
    \midrule
    $R_{15}, R_{45}, R_{60}, R_{75}$ & $R_{0}, R_{30}$  & 89.30  & 88.89  & \textbf{89.53}  \\
    \midrule
    $R_{15}, R_{30}, R_{60}, R_{75}$ & $R_{0}, R_{45}$  & 89.16  & 89.18  & \textbf{89.29}  \\
    \midrule
    $R_{15}, R_{30}, R_{45}, R_{75}$ & $R_{0}, R_{60}$  & 88.72  & 89.10  & \textbf{89.18}  \\
    \midrule
    $R_{15}, R_{30}, R_{45}, R_{60}$ & $R_{0}, R_{75}$  & 84.55  & \textbf{84.64}  & 84.59  \\
    \midrule
    $R_{0}, R_{45}, R_{60}, R_{75}$ & $R_{15}, R_{30}$  & 92.62  & 92.56  & \textbf{92.72}  \\
    \midrule
    $R_{0}, R_{30}, R_{60}, R_{75}$ & $R_{15}, R_{45}$  & 94.17  & \textbf{94.43} & 94.33   \\
    \midrule
    $R_{0}, R_{30}, R_{45}, R_{75}$ & $R_{15}, R_{60}$  & 94.58  & \textbf{94.65}  & 94.61   \\
    \midrule
    $R_{0}, R_{30}, R_{45}, R_{60}$ & $R_{15}, R_{75}$  & 90.16  & 90.16  & \textbf{90.19}  \\
    \midrule
    $R_{0}, R_{15}, R_{60}, R_{75}$ & $R_{30}, R_{45}$  & 92.09  & 92.41  & \textbf{92.62}  \\
    \midrule
    $R_{0}, R_{15}, R_{45}, R_{75}$ & $R_{30}, R_{60}$  & 94.35  & \textbf{94.44}  & 94.40  \\
    \midrule
    $R_{0}, R_{15}, R_{45}, R_{60}$ & $R_{30}, R_{75}$  & 89.93  & 90.13  & \textbf{90.24}  \\
    \midrule
    $R_{0}, R_{15}, R_{30}, R_{75}$ & $R_{45}, R_{60}$  & 92.00  & \textbf{92.28}  & 92.22  \\
    \midrule
    $R_{0}, R_{15}, R_{30}, R_{60}$ & $R_{45}, R_{75}$  & 89.47  & 89.54  & \textbf{89.65}  \\
    \midrule
    $R_{0}, R_{15}, R_{30}, R_{45}$ & $R_{60}, R_{75}$  & 74.14  &  74.80 & \textbf{74.83} \\
    \midrule
    \multicolumn{2}{c}{\textbf{Average}}& 88.31 & 88.45 & \textbf{88.53}\\
    \bottomrule
\end{tabular}}
\end{table}

\section{Conclusion} \label{section:conclusion}
In this paper, we proposed DADG, a novel domain generalization approach, that contains two main components, discriminative adversarial learning and meta-learning based cross domain validation. The discriminative adversarial learning component learns a domain-invariant feature extractor, while the meta-learning based cross domain validation component trains a robust classifier for the objective task (i.e., classification task). 
Extensive experiments have been conducted to show that our feature extractor and classifier could achieve good generalization performance on three domain generalization benchmark datasets. 
Experimental results indicate that the feature extractor and classifier achieve good generalization on three benchmark domain generalization datasets. 
The experimental results also show that our approach consistently beat the strong baseline DeepAll. For instance, while using PACS dataset, our approach performs better than DeepAll by 1.56\% (AlexNet) and 3.11\% (ResNet-18). Notably, we also reach the state-of-the-art performance on VLCS and Office-Home datasets, and improve the average accuracy by over 1\% in each case. 
As we mentioned in the \ref{domain_relation_anay}, in current stage, our DADG is sensitive to the number of source domains because the DAL tends to be unimportant with the increasing number. In the future work, we plan to address this limitation and design a approach that can handle various number of source domains.

\section*{Acknowledgments} Effort sponsored in part by United States Special Operations Command (USSOCOM), under Partnership Intermediary Agreement No. H92222-15-3-0001-01. The U.S. Government is authorized to reproduce and distribute reprints for Government purposes notwithstanding any copyright notation thereon. \footnote{The views and conclusions contained herein are those of the authors and should not be interpreted as necessarily representing the official policies or endorsements, either expressed or implied, of the United States Special Operations Command.}

\bibliography{kdd2020}

\end{document}